\documentclass{article}

\usepackage{arxiv}

\usepackage[utf8]{inputenc} 
\usepackage[T1]{fontenc}    
\usepackage{hyperref}       
\usepackage{url}            
\usepackage{booktabs}       
\usepackage{amsfonts}       
\usepackage{nicefrac}       
\usepackage{microtype}      
\usepackage{cleveref}       
\usepackage{lipsum}         
\usepackage{graphicx}
\usepackage[square,numbers,compress]{natbib}
\usepackage{doi}
\usepackage{color}
\usepackage{listings}

\definecolor{mygreen}{rgb}{0,0.6,0}
\definecolor{mygray}{rgb}{0.5,0.5,0.5}
\definecolor{mymauve}{rgb}{0.58,0,0.82}

\title{Deep-Learning Based Docking Methods: Fair Comparisons to Conventional Docking Workflows}

\date{December 2, 2024}

\newif\ifuniqueAffiliation
\uniqueAffiliationtrue

\ifuniqueAffiliation 
\author{ \href{https://orcid.org/0000-0003-4641-8501}{\includegraphics[scale=0.06]{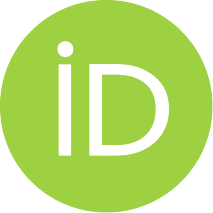}\hspace{1mm}Ajay N.~Jain} \thanks{Corresponding Author} \\
	BioPharmics Division\\
	Optibrium Ltd.\\
	Cambridge, England \\
	\texttt{ajay@optibrium.com} \\
	\And
	\href{https://orcid.org/0000-0002-1622-2770}{\includegraphics[scale=0.06]{orcid.pdf}\hspace{1mm}Ann E.~Cleves} \\
	BioPharmics Division\\
	Optibrium Ltd.\\
	Cambridge, England \\
	\texttt{ann@optibrium.com} \\
	\And
	\href{https://orcid.org/0000-0003-2860-7958}{\includegraphics[scale=0.06]{orcid.pdf}\hspace{1mm}W. Patrick Walters} \\
        Computation and Informatics \\
	Relay Therapeutics\\
	Cambridge, MA, USA \\
	\texttt{pwalters@relaytx.com} \\
}
\else
\usepackage{authblk}

\newbox{\orcid}\sbox{\orcid}{\includegraphics[scale=0.06]{orcid.pdf}} 
\author[1]{%
	\href{https://orcid.org/0000-0003-4641-8501}{\usebox{\orcid}\hspace{1mm}Ajay N.~Jain\thanks{\texttt{ajay@optibrium.com}}}%
}
\author[1,2]{%
	\href{https://orcid.org/0000-0002-1622-2770}{\usebox{\orcid}\hspace{1mm}Ann E.~Cleves\thanks{\texttt{ann@optibrium.com}}}%
}
\affil[1]{Department of Computer Science, Cranberry-Lemon University, Pittsburgh, PA 15213}
\affil[2]{Department of Electrical Engineering, Mount-Sheikh University, Santa Narimana, Levand}
\fi

\renewcommand{\headeright}{Brief Perspective}
\renewcommand{\undertitle}{Brief Perspective}
\renewcommand{\shorttitle}{Conventional Docking vs. Deep Learning}

\hypersetup{
pdftitle={Fair Comparisons with DiffDock},
pdfsubject={q-bio.NC, q-bio.QM},
pdfauthor={Ajay N.~Jain, Ann E.~Cleves},
pdfkeywords={Surflex-Dock, Cognate Docking, Glide, AutoDock Vina, Gnina},
}

\begin{document}
\maketitle

\begin{abstract}
The diffusion learning method, DiffDock, for docking small-molecule ligands into protein binding sites was recently introduced. Results included comparisons to a number of more conventional docking approaches, with DiffDock showing nominally much better performance. Here, we employ a fully automatic workflow using the Surflex-Dock methods to generate a fair baseline for conventional docking approaches. Results were generated for the common and expected situation where a binding site location is known and also for the condition where the entire protein was the nominal target of docking. For the known binding site condition, Surflex-Dock success rates at 2.0\AA~RMSD far exceeded those for DiffDock (Top-1/Top-5 success rates, respectively, were 68/81\% compared with 45/51\%). Glide performed with similar success rates (67/73\%) to Surflex-Dock for the known binding site condition, and results for AutoDock Vina and Gnina followed this pattern. For the unknown binding site condition, using an automated method to identify multiple binding pockets, Surflex-Dock success rates again exceeded those of DiffDock, but by a somewhat lesser margin. DiffDock made use of roughly 17,000 co-crystal structures for learning (98\% of PDBBind version 2020, pre-2019 structures) for a training set in order to predict on 363 test cases (2\% of PDBBind 2020) from 2019 forward. The Surflex-Dock approach made \emph{no use} of prior information from pre-2019 structures, either for binding site identification or for knowledge to guide the docking and pose-ranking process. DiffDock's performance was inextricably linked with the presence of near-neighbor cases of close to identical protein-ligand complexes in the training set for over half of the test set cases. DiffDock exhibited a roughly 40 percentage point difference on near-neighbor cases (two-thirds of all test cases) compared with cases for which a near-neighbor training case was not identified. DiffDock has apparently encoded a type of table-lookup during its learning process, rendering any meaningful application scenario beyond its reach. Further, it does not perform even close to competitively with a competently run modern docking workflow.
\end{abstract}


\section{Introduction}
Deep learning approaches have recently generated significant interest in the field of computer-aided drug design (CADD), and this phenomenon is exemplified by the excitement produced by the introduction of the DiffDock method \citep{diffdock}. This short report is not intended to review the field or to differentiate substantive and meaningful deep-learning contributions to CADD from lesser contributions. Rather, the focus is to make the general CADD community aware of the proper context in which to interpret DiffDock's recently reported results. Our work should be considered alongside another recent contribution, a paper from the  Deane group, which offers additional context for understanding the results of AI-based docking methods like DiffDock, highlighting, in particular, extremely strained conformations in predicted docked poses \citep{posebusters}.

Briefly, the performance of DiffDock was demonstrated by training on 98\% of the co-crystal structure data from PDBBind version 2020 (pre-2019 structures) and testing on the remaining 2\% (2019 forward). In what follows, we will lay out two main points. First, a mature and fully automated docking workflow far outperforms DiffDock on its own test set \emph{without} using any prior knowledge of other co-crystal structures. This is true whether the ligand's binding site location is provided to the docking system or not (so-called ``blind docking''). Second, near-neighbor training cases, where both the protein binding site \emph{and} the cognate ligand are nearly identical to those from a test case, exist for the majority of DiffDock's PDBBind testing set. DiffDock performance on these near-neighbor test cases was \emph{much} better than on non-near-neighbor cases.

\begin{figure}[!b]
	\centering
        \includegraphics[scale=0.8]{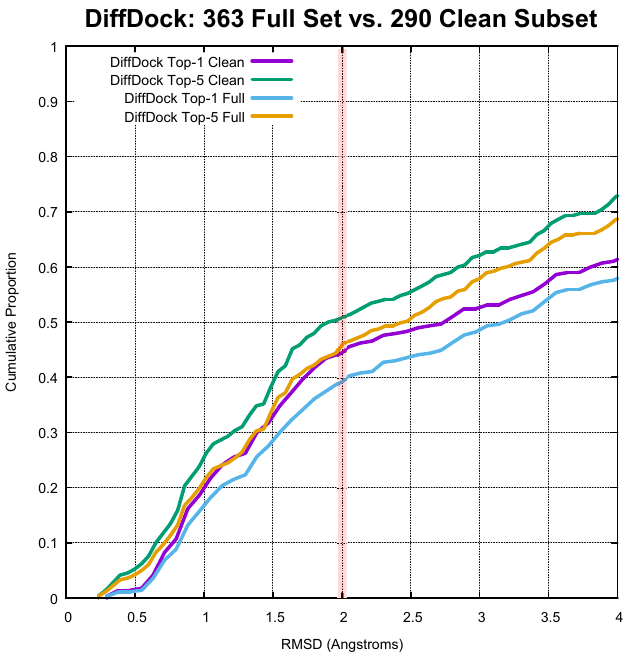}
\caption{Cumulative histogram of RMSD values for Top-1 and Top-5 DiffDock predicted poses for the Full Test Set (cyan and yellow curves) and the Clean Test Set from full PDB re-processing (violet and green curves).}
	\label{clean-set}
\end{figure}

\section{Results}
Details of the specific computational procedures will be presented in the Data and Methods Section (with additional details in the Appendices), below. The goals of the work presented here were two-fold: 1) to establish a reasonable baseline for the performance of conventional docking workflows on the DiffDock test set; and 2) to try to understand the underlying performance driver for DiffDock's apparent success.

\subsection{DiffDock Performance: As Reported compared with a ``Clean'' Test Subset}
The PDBBind 2020 data set was used to train and test DiffDock, making use of roughly 17,000 protein-ligand complexes for training and 363 complexes for testing (the ``Full Test Set''). Here, we made use of a fully automated pipeline to process the PDB complexes, following previously reported protocols \citep{pinc2015}. The processed PDB complexes consisted of protein and ligand files, with bond orders assigned and with protonation as expected in physiological pH. Ligand structures were subjected to several quality tests, including cross-checking bond-order assignments against curated SMILES representations of corresponding the PDB HET codes. Those cases where quality tests were passed and where the resulting ligand structures topologically agreed with those curated for the DiffDock benchmarking were kept as a ``clean'' set for testing conventional docking workflows (the ``Clean Test Set''). There were 290 such cases (80\% of the total 363 original test cases).

Figure \ref{clean-set} shows the performance of DiffDock on its Full and Clean Test Sets. The cyan and violet curves correspond to cumulative histograms of the RMS deviations for the the top-scoring pose for the full and clean test sets, respectively. The yellow and green violet curves correspond to cumulative histograms of the RMS deviations for the the best of the top five poses for the full and clean test sets, respectively. There was a relatively marginal improvement on the Clean 290 complex subset, amounting to an increase of roughly five percentage points in success rates at the 2.0\AA~RMSD threshold, which \emph{favors} the DiffDock method compared with the original published Full Test Set. This was presumably due to the automatic quality checks that the \texttt{grindpdb} methodology employs (see Data and Methods) and also because two independent curation approaches agreed on the bound ligand structures. In what follows, comparative results will be presented on the Clean Test Set. 

\subsection{DiffDock Performance: Comparison to Conventional Docking Workflows}
The original report of DiffDock made comparisons to a number of conventional docking methods, but the methods were employed in an unconventional manner. Rather than providing an explicitly scoped binding site, a so-called ``blind'' docking procedure was used where docking was performed against entire protein structures. Consequently, whereas mature docking methods in cognate ligand re-docking typically perform with roughly 60--80\% success at the 2.0\AA~RMSD threshold \citep{spitzer2012} for top-scoring poses, the DiffDock reported performance for GNINA, SMINA, and Glide ranged from 19--23\% \citep{diffdock}. In what follows, we will present comparative results for Surflex-Dock \citep{jain2003,jain2007,jain2009,pinc2015}, Glide \citep{Friesner2004,Halgren2004}, AutoDock Vina \citep{vina2010,vina2021}, and Gnina \citep{gnina2021} using conventional cognate-ligand re-docking with a defined binding site. Results for Surflex-Dock, which has a well-studied pocket finding algorithm \citep{spitzer2013,Spitzer2014}, will also be presented in the unknown binding-site condition.

\begin{figure}[!b]
	\centering
        \includegraphics[scale=0.8]{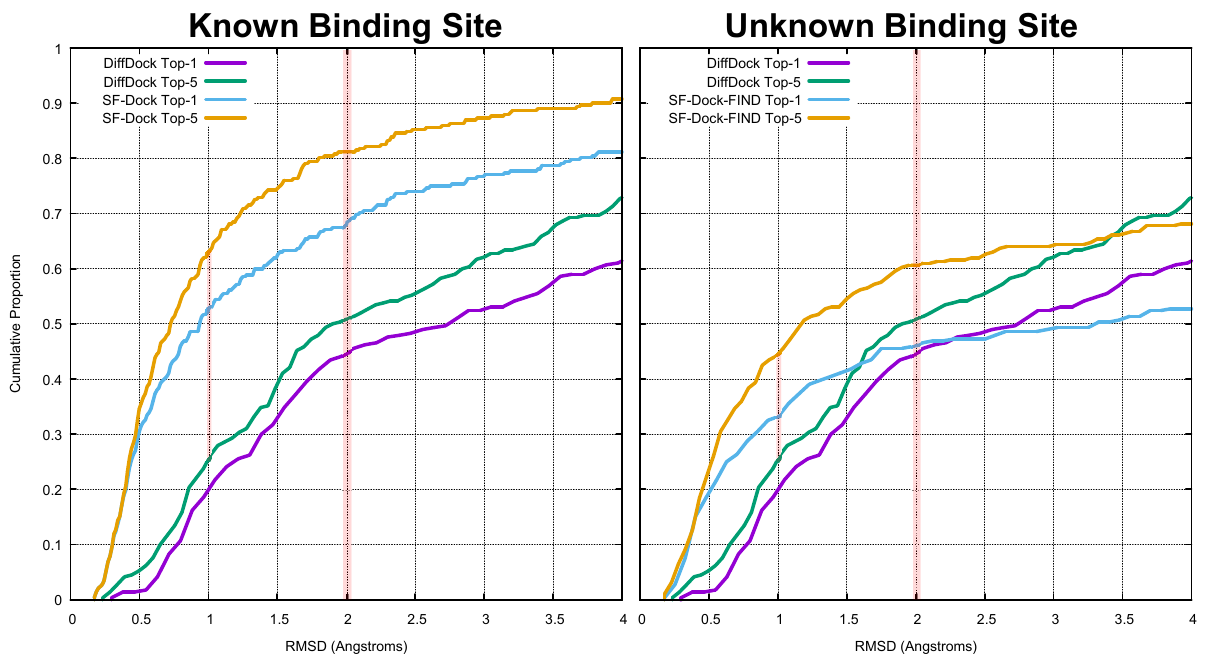}
\caption{Cumulative histogram of RMSD values for Top-1 and Top-5 Surflex-Dock (cyan and yellow) and DiffDock poses (violet and green) for the Clean Test Set with known binding sites (left) and with unknown binding sites (right).}
	\label{sf-dock}
\end{figure}

\subsubsection{Comparison to Surflex-Dock}
Figure \ref{sf-dock} shows the comparison between pose prediction accuracy for two different testing conditions. At left, Surflex-Dock was run using a conventional docking protocol where the location of the binding site was known. Cumulative histograms of RMSD are shown for DiffDock (violet and green curves) and for Surflex-Dock (cyan and yellow curves). Focusing on the 2\AA~RMSD threshold, we see a roughly 25--30 percentage point advantage for Surflex-Dock, and an even larger gap at the 1.0\AA~threshold. The performance difference is both practically and statistically highly significant (p $< 10^{-10}$ by paired t-test for both Top-1 and Top-5 pose prediction performance).

At right, Surflex-Dock was run using a protocol that lacked a defined binding site, requiring the use of an automated procedure to identify possible binding sites on each protein, docking to up to ten of the highest ranked candidates, and combining results based purely on docking scores. At the 1.0\AA~threshold, in the unknown binding site condition (so-called  ``blind docking''), results favored Surflex-Dock by 15--20 percentage points. At 2.0\AA, top-scoring pose performance was roughly equivalent but among the top five poses, Surflex-Dock had a roughly 10 percentage point advantage. Statistical comparison of performance on the full 290 complexes was complicated by the presence of extreme outliers for each method, where an incorrect binding site scored better than the correct one. For the 160 cases where both methods had RMSD in the top five condition of 4.0\AA or better (the subset seen in the plot of Figure \ref{sf-dock}), Surflex-Dock was statistically superior to DiffDock by paired t-test (p $< 0.01$ for the top scoring poses and p $< 10^{-8}$ for the top five).

It is important to contextualize the results in the unknown binding site condition. The DiffDock procedure relied on roughly 98\% of the PDBBind 2020 set for training, which explicitly identified, in many cases, the known binding site for proteins either identical to or similar to those in the test set. Given, for example, a number of estrogen-receptor protein-ligand complexes as prior knowledge, the binding site is well-defined for a test complex of estrogen-receptor for a new ligand. The procedure used for Surflex-Dock in the ``blind docking'' condition used \emph{no prior knowledge} of any kind.

\subsubsection{Comparison to Glide}
In the original DiffDock report, the authors made use of a procedure to run the Glide docking method on the entire protein, rather than making use of a defined binding site, which is the use case for which the method has been developed and optimized. Consequently, the reported results (roughly 22\% success at 2.0\AA~RMSD) were at variance with expectations for cognate-ligand re-docking (closer to 60--80\% success \citep{spitzer2012}).

\begin{figure}[!b]
	\centering
        \includegraphics[scale=0.8]{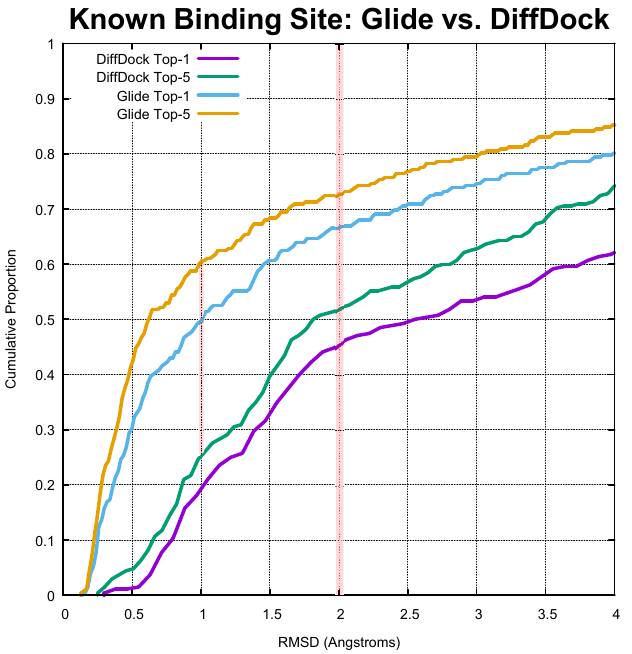}
\caption{Cumulative histogram of RMSD values for Top-1 and Top-5 Glide (cyan and yellow) and DiffDock poses (violet and green) for the Clean Test Set with known binding sites.}
	\label{glide}
\end{figure}

Here, we report Glide results for 272 complexes whose proteins were correctly processed using an automatic preparation procedure (see Data and Methods). Of these, 260/272 yielded successful dockings, with the remaining 12 being assigned values of 20.0\AA~for Top-1 and Top-5 pose accuracy. Figure \ref{glide} shows the comparison between pose prediction accuracy for the known binding-site condition. Cumulative histograms of RMSD are shown for DiffDock (violet and green curves) and for Glide (cyan and yellow curves). Focusing on the 2\AA~RMSD threshold, we see a roughly 25 percentage point advantage for Glide, and an even larger gap at the 1.0\AA~threshold. The performance difference is both practically and statistically highly significant (p $< 10^{-7}$ by paired t-test for Top-1 pose prediction and p $= 10^{-5}$ for Top-5).

\subsubsection{Comparison to AutoDock Vina and Gnina}
The DiffDock report presented results for Smina and Gnina \citep{smina2013,gnina2021}, both of which were derived from AutoDock Vina \citep{vina2010,vina2021}, a widely used open-source method. Gnina uses a form of deep learning to improve Vina performance, both for pose prediction and for virtual screening applications \citep{gnina2021}. Here, we report ligand re-docking results for both Vina and Gnina using the known binding-site condition. Of the full Clean Test Set of 290 complexes, 285 were processed correctly using the standard AutoDock procedures (see Data and Methods) to yield usable protein structures. For both Vina and Gnina, all 285 complexes yielded docking results, with no failures.

\begin{figure*}[!b]
	\centering
        \includegraphics[scale=0.8]{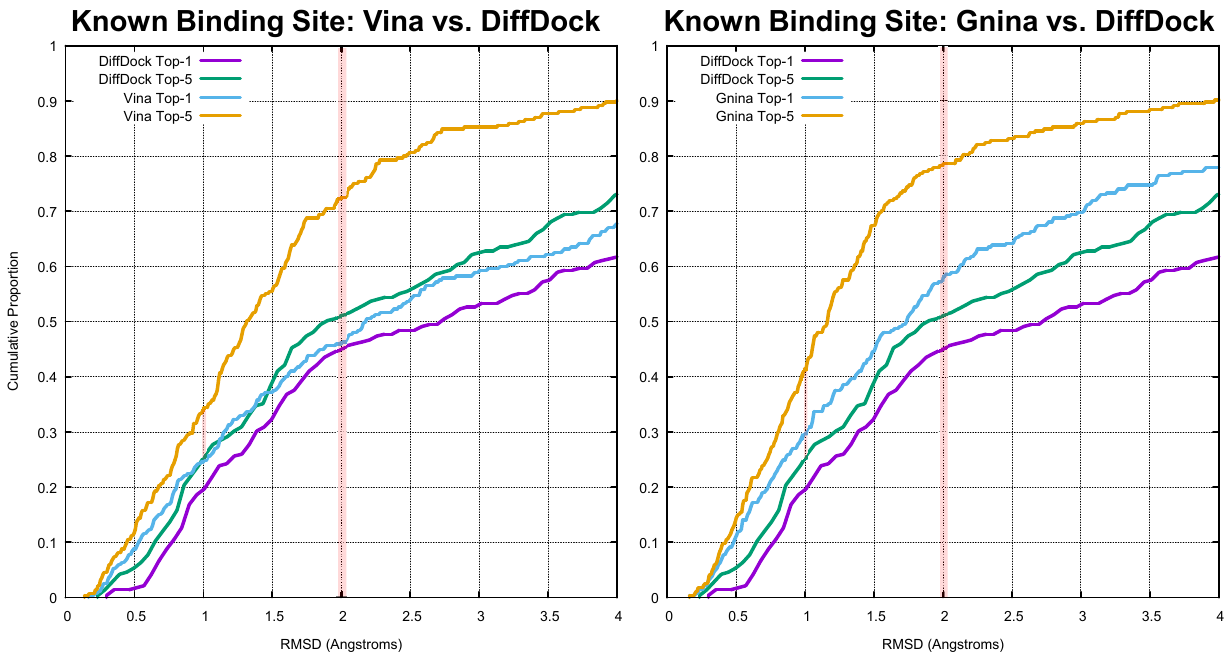}
\caption{Cumulative histograms of RMSD values for Top-1 and Top-5 Vina and Gnina (cyan and yellow, left and right, respectively) and DiffDock poses (violet and green) for the Clean Test Set with known binding sites.}
	\label{vina}
\end{figure*}

Figure \ref{vina} shows the comparison between pose prediction accuracy for the known binding-site condition. Cumulative histograms of RMSD are shown for DiffDock (violet and green curves) and for Vina (left) and Gnina (right), with cyan and yellow curves to depict Top-1 and Top-5 pose prediction performance, respectively. Focusing on the 2\AA~RMSD threshold, the results for both Vina and Gnina exceeded those of DiffDock by roughly 20--25 percentage points. As previously reported \citep{gnina2021}, the Gnina method exhibited improved performance for top-ranked pose (right, cyan curve) over both DiffDock (purple curve) and Vina (cyan curve, left-hand plot). Note that neither method approached the success levels of either Glide or Surflex-Dock at the 1.0\AA~RMSD threshold. As with the previous comparisons, the performance differences, particularly between DiffDock and Gnina, were both practically and statistically highly significant. For Vina Top-1 and Top-5 pose prediction, respectively, p values by paired t-test were $< 10^{-7}$ and p $= 10^{-9}$ for Top-5). For Gnina, the p values were less than $10^{-10}$ in both cases.

Note, however, that results for Gnina are difficult to interpret, due to its reliance on extensive training data for its scoring method, which implements a convolutional neural-network that was trained on hundreds of protein-ligand complexes. Rather than this training resulting in a scoring function, as is used by Surflex-Dock, Glide, and Vina, this approach essentially learns characteristics of ``native-like'' predicted ligand poses from explicit training data. It is possible that the internal representation induced by the Gnina scoring function has, to some extent, memorized binding motifs that are directly represented in the DiffDock test set. This type of effect will be explored directly, with respect to DiffDock performance in what follows.

\subsection{DiffDock Performance: Effects of Near-Neighbor Training Cases}
Recall that DiffDock was trained using a temporal 98/2\% split of data, with roughly 17,000 PDB complexes serving as training and in the Clean Set, just under 300 structures for testing. Given the power of deep learning methods to build complex internal representations, it is possible that DiffDock essentially ``memorized'' many training structures that were helpful for some subset of complexes used for testing.

\begin{figure}[!t]
	\centering
        \includegraphics[scale=0.7]{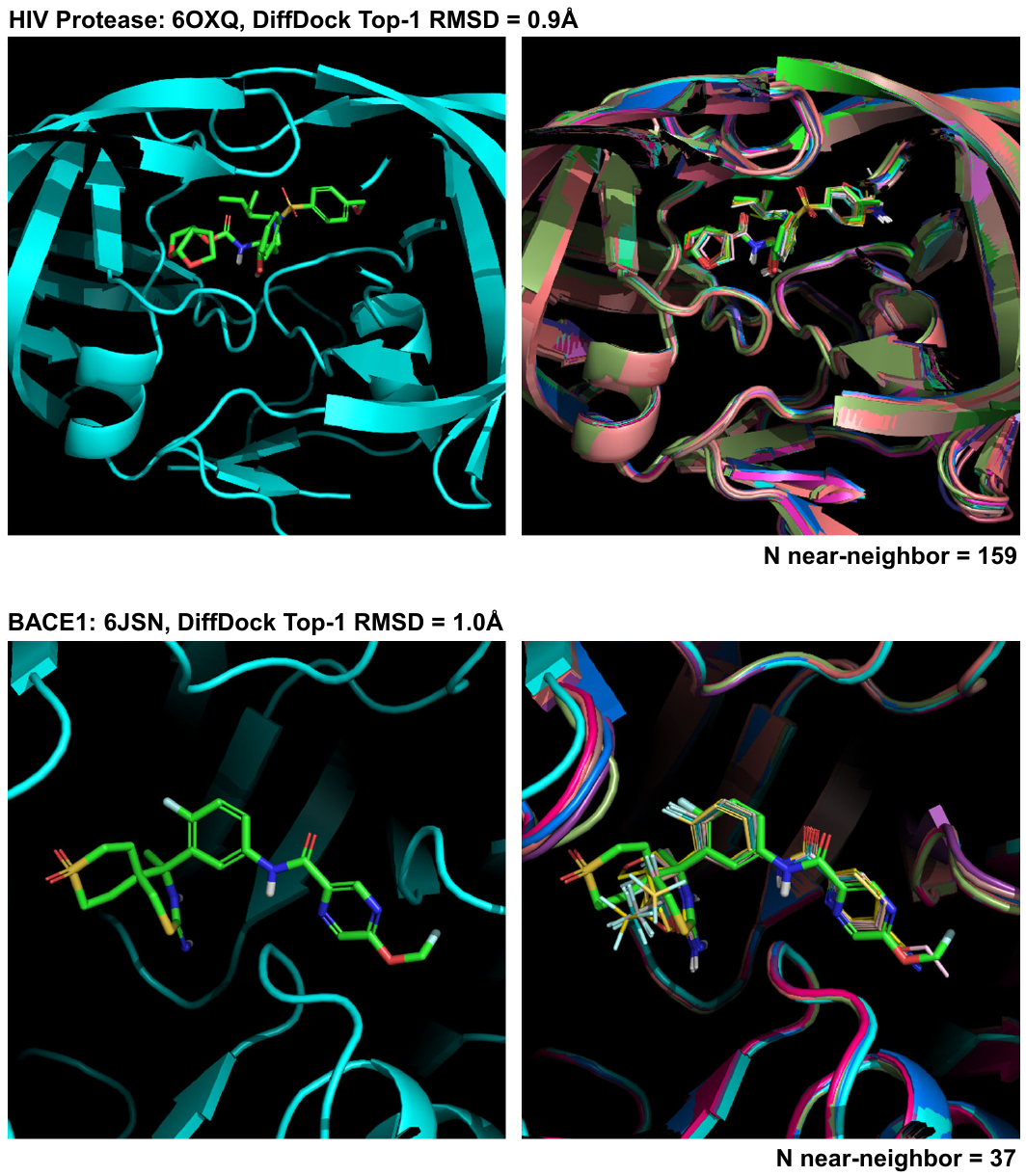}
\caption{Examples of near-neighbor success cases for DiffDock: HIV-protease (top) and BACE1 (bottom).}
	\label{nn1}
\end{figure}

Figure \ref{nn1} shows two test cases on which DiffDock ``predicted'' well, with bound inhibitors of HIV protease (top) and BACE1 (bottom). At left, we show the test complexes with their cognate ligands, whose poses are to be predicted. At right, we see an alignment of \emph{training} data cases that had extremely high binding pocket similarity \emph{and} 2D ligand similarity to the ligands to be predicted. For HIV Protease, the number of near-neighbor cases based on binding pocket similarity was 159, with a subset differing by only a few atoms from the test ligand of 6OXQ. For BACE1, the number of near-neighbor cases based on binding pocket similarity was 37, again with a subset differing by only a few atoms from the test ligand of 6JSN.

\begin{figure}[!t]
	\centering
        \includegraphics[scale=0.7]{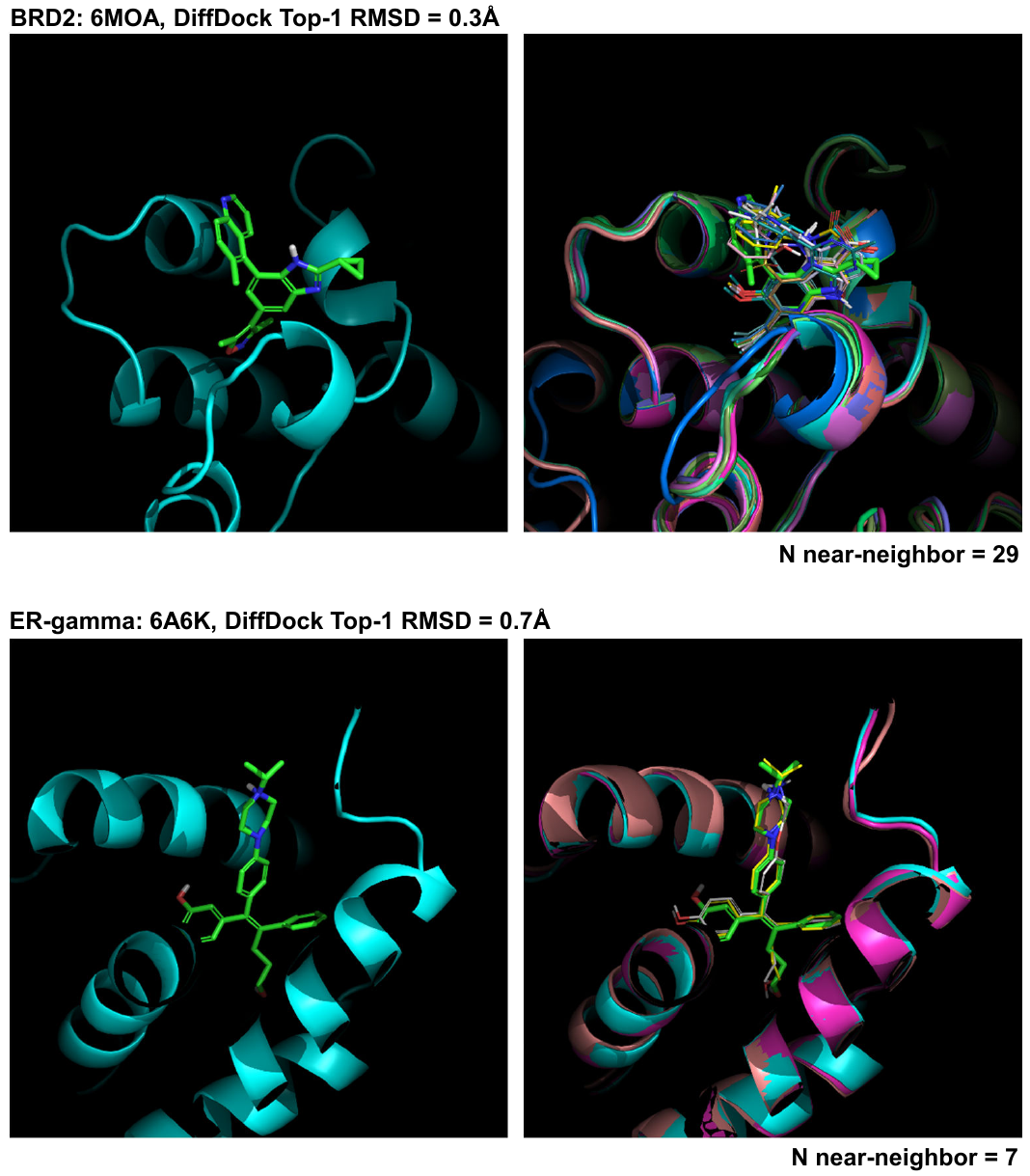}
\caption{Examples of near-neighbor success cases for DiffDock: BRD2 (top) and ER-$\gamma$ (bottom).}
	\label{nn2}
\end{figure}

Figure \ref{nn2} shows two additional test cases on which DiffDock ``predicted'' well, with bound ligands of BRD2 (top) and ER-$\gamma$ (bottom). At left, we show the test complexes with their cognate ligands, whose poses are to be predicted. At right, we see an alignment of \emph{training} data cases that had extremely high binding pocket similarity \emph{and} 2D ligand similarity to the ligands to be predicted. For BRD2, the number of near-neighbor cases based on binding pocket similarity was 29, with a subset differing by only a few atoms from the test ligand of 6MOA. For ER-$\gamma$, the number of near-neighbor cases based on binding pocket similarity was 7, again with a subset differing by only a few atoms from the test ligand of 6A6K.

Clearly, having a database of examples that contain nearly identical binding sites and ligands solves \emph{both} the ``prediction'' problem of where a protein's binding site is \emph{and} almost exactly how the ligand binds. So how large was the effect of near-neighbor training set cases on DiffDock's reported performance?

We performed a systematic analysis to identify those Clean Test Set cases that had near neighbors within the DiffDock training set, defined as having both high topological similarity between the test and training ligands \emph{and} having high protein binding site similarity (see Appendices for additional details on the computational procedures). Overall, 191/290 test cases (roughly two-thirds) had a near-neighbor training case and, of these, 24 were termed extreme near-neighbors based on very high topological ligand similarity. Just 99/290 (roughly one-third) of test cases had no identified near-neighbor training cases using this non-exhaustive procedure.

\begin{figure}[!t]
	\centering
        \includegraphics[scale=0.9]{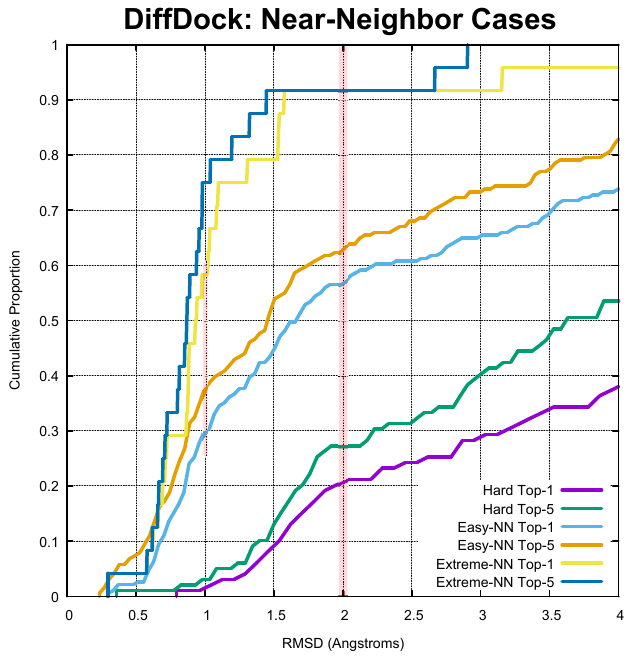}
\caption{Cumulative histogram of RMSD values for Top-1 and Top-5 DiffDock predicted poses for three subsets: hard cases with no identified near-neighbors (violet and green), standard near-neighbor cases (cyan and yellow), and extreme near-neighbor cases (yellow and blue).}
	\label{nn-plot}
\end{figure}

Figure \ref{nn-plot} shows the performance separation of different sets of cases, grouped by their relative challenge in terms of the existence of near-neighbor training cases. Overall, for the 191/290 cases that fell into the near-neighbor set (cyan and orange curves), success rates were 57/65\% for Top-1/Top-5, respectively. Results for the 99/290 more challenging cases (violet and green) showed success rates of roughly 21/28\% for Top-1/Top-5, respectively. The set of 24 extreme near neighbors (yellow and blue curves), by contrast, exhibited over 90\% success for Top-1 and Top-5 pose predictions. 

The original DiffDock report's success rates for the Clean Test Set of 290 complexes were 45/51\% for Top-1/Top-5, respectively. Clearly, the apparent level of success was greatly influenced by the presence of a subset of nearly two-thirds of cases where simple table-lookup could produce a good answer, and where DiffDock performed nominally well (57/65\%). Note that even in the subset of near-neighbor cases (cyan and orange curves in Figure \ref{nn-plot}), DiffDock's performance was inferior to competently run, fully automatic docking from the mature commercial methods Surflex-Dock and Glide, whose success rates were 67--68/73--81\% (Top-1/Top-5) on the complete Clean Test Set. AutoDock Vina produced performance of roughly 47/73\%, clearly better with respect to Top-5 pose prediction performance. Gnina, with the caveat of potentially the same type of train/test set contamination suffered by DiffDock, performed substantially better than DiffDock on the near-neighbor set (58/78\%).

On the roughly one-third of cases where automated procedures did not find near neighbors, DiffDock's performance was \emph{less than half} as good as on the two-thirds of near-neighbor cases. DiffDock's success or failure was dichotomized by the presence or absence of near-neighbors in the training set: essentially a form of table-lookup.

\section{Data and Methods}
All data was derived from the PDBBind version 2020 database, which also served as the primary source for DiffDock validation using experimentally determined crystal structures. All computational procedures employing the BioPharmics Platform software made use of version 5.173 unless otherwise noted (BioPharmics Division, Optibrium Limited, Cambridge, CB25 9GL, UK). Modules used included the Tools, Docking, and Similarity modules \citep{cleves2017,fgen2019,jain2003,cleves2020,pinc2015,spitzer2013,esim2019}. This included procedures for automatic PDB file processing into proteins and ligands, randomization of ligand conformations, docking, calculation of automorph-corrected RMSD, and ligand/protein similarity calculations to identify near-neighbor training set mates to the Clean Test Set. For Glide docking \citep{Friesner2004,Halgren2004}, the Schr{\"o}dinger 2022-3 Suite was employed (Schr{\"o}dinger LLC, New York, NY, 2023). For docking with Vina \citep{vina2010,vina2021}, version 1.1.2 was used, and for Gnina \citep{gnina2021}, version 1.1 (master:e4cb300+, Built 12-18-2023).

Please refer to the Appendices for additional details.

\section{Conclusions}
It is possible that the DiffDock training procedure has learned an interesting encoding of a large set of structures of protein-ligand complexes. However, what it appears to be doing cannot be considered to be either ``docking'' or ``identification'' of ligand binding sites. The reported results are overwhelmingly contaminated with near neighbors to test cases within the extensive training set.

Further, results reported for mature and widely used docking methods presented an extremely misleading baseline of comparison. Surflex-Dock, Glide, Vina, and Gnina \emph{all} performed much better than DiffDock on cognate ligand re-docking in the known binding-site condition. Surflex-Dock, which has a mature and automatic method to identify binding sites, also performed much better than DiffDock in the unknown binding-site condition.

The primary reported results for DiffDock were artifactual, and the comparative results for other methods were incorrectly done. We do not mean to suggest that the study's authors were mendacious in any way. The intention of our report is to be constructive in offering an instructive and carefully done analysis.

There can be value gained from different disciplines applying powerful general methods to new domains. With respect to machine-learning and computer-aided drug design, we would like to offer some observations to consider prior to making strong claims of superiority over pre-existing methods:

\begin{itemize}

\item The CADD field has a long history, and the significant and challenging problems have changed over time. Cognate ligand re-docking has not been an important problem for well over a decade. Predicting bound poses of ligands with \emph{novel structures} compared to prior known compounds, and into (obviously) \emph{non-cognate protein binding sites}, is a challenging and important problem. It is important for newcomers to CADD to understand what has been done before and the problems that will make a true difference to the field if solved. It is easy (and tempting) to unintentionally develop artificial benchmarks that do not reflect the ultimate application of a method.

\item Small molecules are generally not experiments of nature, but are typically designed by people who have biases both with respect to which targets are thought to be interesting and, more importantly, with respect to prior known ligands. So, very often, a compound made and co-crystallized with the protein for which it was designed to bind will have extremely similar prior analogs. Temporal train/test splitting, as was done in the DiffDock study, is the correct idea, but the split cannot be 98/2\%, else simple memorization of the training set can dominate results. A more reasonable temporal split for non-cognate docking is 25/75\% \citep{pinc2015}.

\item Generally speaking, CADD methods are complex, particularly for docking, where aspects of protein preparation and binding site definition exist in addition to the challenges in appropriately representing and manipulating small molecules. Running an unfamilar method according to reasonable practices can be tricky and needs to be done respecting the designed application scenario. When results are obtained that are at large variance to prior published work, as in the DiffDock report with respect to Glide's performance on cognate ligand re-docking, care should be taken to understand what may have gone wrong.

\item CADD is not like speech recognition or optical character recognition, where correctly predicting new examples that are very close to prior known data, and which are drawn from the same population as the known data, is the main use-case scenario. The goal in CADD is to make compounds that have \emph{different} properties, often with novel scaffolds, than prior known compounds for a target or to design compounds to modulate the activity of a new target. There are areas where predictions on subtle changes to a parent compound are important, for example in affinity prediction, metabolism, toxicity, etc... Pose prediction for minor variants on a parent compound is generally not difficult or challenging, except in cases where a small structural change leads to a large difference in bound pose. But that is exactly the case where ML approaches like DiffDock \emph{will not} work.

\item The most interesting and challenging problems in CADD arise when data are sparse, not when many thousands of relevant data points exist. Methods in ML that rely on large data sets have interesting, successful, and impactful applications (e.g. learned force-fields \citep{ani2017}), but care must be taken to identify problem areas where the data requirements match the application scenario.
    
\end{itemize}

Publication of studies such as the DiffDock report \citep{diffdock} are not cost-free to the CADD field. Magical sounding claims generate interest and take time for groups to investigate and debunk. Many groups must independently test and understand the validity of such claims. This is because most groups, certainly those focused primarily on developing new drugs, do not have the time to publish extensive rebuttals such as this. Therefore their effort in validation/debunking is replicated many fold. The waste of time and effort is substantial, and the process of drug discovery is difficult enough without additional unnecessary challenges.

One cannot make a blanket recommendation to simply ignore all reports from CADD newcomers that make magic claims, because there may be great value in rare cases. However, we can make the strong recommendation that newcomers to CADD heed the foregoing list of five observations.

\section{Addendum}
The analyses reported here were based on the original DiffDock report \citep{diffdock}, with performance data provided directly by authors of that report, corresponding exactly to the published figures and tables. Subsequently, in February 2024, a new benchmark (DockGen) and a new DiffDock version (DiffDock-L) was released by the DiffDock group \citep{diffdock-l}. This work post-dated our analyses, and we were unaware of this work at the time of our initial report, whose release was delayed following completion of the analyses.

\newpage
\section*{Appendix: Surflex-Dock Protein Preparation and Docking Scripts}
All results reported for Surflex-Dock was done using the BioPharmics Surflex Platform software version 5.173 (BioPharmics Division, Optibrium Limited, Cambridge, CB25 9GL, UK). Given a PDB code, the following script (\texttt{RunGrind}) fetches the protein structure from the RCSB PDB, separates protein, ligands, and water molecules, protonates and selects protomer and tautomer states for the protein and ligands:

\begin{lstlisting}
#!/bin/bash
# Script Name: RunGrind
# Arg1 = PDB code
# Note: The PDBBind_processed folder contains individual PDB-named directories

cd PDBBind_processed/$1
echo $1 > PDBList
sf-dock.exe -verifypdb ../verifypdb.smi getpdb PDBList trg
source trg-script
cd ../..

# Example: RunGrind 6a6k
# Key output files: trg-pro-6a6k.mol2  trg-lig-6a6k-*.mol2
\end{lstlisting}
Following the raw processing of a particular PDB code from within the PDBBind set, identification of the ligand to be docked along with cross-verification of its structure was done as follows (\texttt{RunPrep}):

\begin{lstlisting}
#!/bin/bash

# Arg1 = PDB code

cd PDBBind_processed/$1

# Gather all of the parsed ligands into a single mol2 file
cat trg-lig-*.mol2 > all-lig.mol2
# Compare their structures in 2D against the curated PDBBind ligand
sf-sim.exe gsim_list all-lig.mol2 $1_ligand.mol2 gsim

# If a ligand has a 2D gsim score of 1, then we found a corresponding
# ligand to the PDBBind one. Copy it to gold-lig.mol2:
grep "max 1" gsim | head -1 | awk '{print $1}' | sed s/:// | 
      awk '{print "cp trg-lig-" $1 ".mol2 gold-lig.mol2"}' > copy-lig
source copy-lig

# Copy the protein to protein.mol2
cp trg-pro-$1.mol2 protein.mol2

# Build the sf-dock protomol for docking:
echo protein.mol2 gold-lig.mol2 > SiteList
sf-dock.exe mproto SiteList p1

# Randomize the ligand and perform conformer search
# on the agnostic starting conformer
sf-tools.exe regen3d gold-lig.mol2 lig
sf-tools.exe -pgeom forcegen lig-random.mol2 pglig

cd ../..
\end{lstlisting}
As described above, this resulted in 290 clean complexes from the full set of DiffDock's 363 complex test set. The same procedure resulted in 15,268 clean cases from the DiffDock overall set of roughly 17,000 protein-ligand complexes.

\newpage
Cognate ligand re-docking was run in an automated fashion, as follows (\texttt{RunDock}):
\begin{lstlisting}
#!/bin/bash
# Arg1 = PDB code

cd PDBBind_processed/$1

# Run standard -pgeom level docking to the binding site
sf-dock.exe -pgeom gdock_list pglig.sfdb p1-targets pglog

# Cluster poses into pose families
sf-dock.exe posefam pglog

# Measure automorph-corrected RMSD from the PDB deposited bound ligand coordinates
# for the top result and for the top 5 results.
sf-dock.exe rms_fam gold-lig.mol2 pglog rms01 1
sf-dock.exe rms_fam gold-lig.mol2 pglog rms05 5

cd ../..
\end{lstlisting}

Docking to proteins without an identified binding site was done as follows (\texttt{RunFindDock}):
\begin{lstlisting}
#!/bin/bash
# Arg1 = PDB code

cd PDBBind_processed/$1

# Find protein cavities
sf-dock.exe psim_findcav protein.mol2 find

# Dock to all such cavities, for which protomols were generated in the
# prior step
sf-dock.exe -pgeom gdock_list pglig.sfdb ../find00-targets pglogfind00
sf-dock.exe -pgeom gdock_list pglig.sfdb ../find01-targets pglogfind01
sf-dock.exe -pgeom gdock_list pglig.sfdb ../find02-targets pglogfind02
sf-dock.exe -pgeom gdock_list pglig.sfdb ../find03-targets pglogfind03
sf-dock.exe -pgeom gdock_list pglig.sfdb ../find04-targets pglogfind04
sf-dock.exe -pgeom gdock_list pglig.sfdb ../find05-targets pglogfind05
sf-dock.exe -pgeom gdock_list pglig.sfdb ../find06-targets pglogfind06
sf-dock.exe -pgeom gdock_list pglig.sfdb ../find07-targets pglogfind07
sf-dock.exe -pgeom gdock_list pglig.sfdb ../find08-targets pglogfind08
sf-dock.exe -pgeom gdock_list pglig.sfdb ../find09-targets pglogfind09

# A little kludge to deal with the multiple runs for pose family finding
# We need to concatenate the poses for all dockings before building pose families.
cp pglogfind00 pglogfindall
cat pglogfind01 pglogfind02 pglogfind03 pglogfind04 pglogfind05 
    pglogfind06 pglogfind07 pglogfind08 pglogfind09 | grep -v atom >> pglogfindall
cat pglogfind00-results.mol2 pglogfind01-results.mol2 pglogfind02-results.mol2 
    pglogfind03-results.mol2 pglogfind04-results.mol2 > pglogfindall-results.mol2
cat pglogfind05-results.mol2 pglogfind06-results.mol2 pglogfind07-results.mol2
    pglogfind08-results.mol2 pglogfind09-results.mol2 >> pglogfindall-results.mol2

# Generate pose families, but use pose numbers rather than pose names    
sf-dock.exe +pf_mnum posefam pglogfindall

# Measure automorph-corrected RMSD from the PDB deposited bound ligand coordinates
# for the top result and for the top 5 results.
sf-dock.exe rms_fam gold-lig.mol2 pglogfindall rmsfind01 1
sf-dock.exe rms_fam gold-lig.mol2 pglogfindall rmsfind05 5

cd ../..

\end{lstlisting}
\newpage
\section*{Appendix: Finding Near-Neighbor Training Cases}
The following script was run within on Clean Test Set case, looking for any training set ligands that we highly similar to the cognate test ligand. Those training cases with cognate ligands within the top 1\% of Gsim values or whose Gsim value was 0.3 or greater were then aligned to the test case binding site using the PSIM method. All cases where the PSIM score was $\geq 0.65$ and where the training ligand was non-identical to the test ligand were considered ``easy'' cases (191 total). Of those, when the Gsim value was  $\geq 0.80$, these were termed extreme near neighbors (24 of the 191 near-neighbor easy cases).

\begin{lstlisting}
###################################################################################
#!/bin/bash
export OMP_THREAD_LIMIT=8

# Arg1 = PDB code

cd PDBBind_processed/$1

sf-sim.exe gsim_list ../AllLigs.sfdb gold-lig.mol2 gsim-log

# Top 1% of Gsim results roughly 160 mols OR Gsim >= 0.30

cat gsim-log | sort -nr -k6 | awk '{if (($6 >= 0.30) || (NR <= 160)) print $1}' | sed s/:// | sed s/-/\ / | awk '{print "../" $1 "/protein.mol2 ../" $1 "/gold-lig.mol2"}' > plistgsim100

sf-dock.exe +psim_dump psim_list plistgsim100 protein.mol2 gold-lig.mol2 proteinaligngsim100

cd ../..
###################################################################################

###################################################################################
# 191 near-neighbor easy cases:
5zr3 5zxk 6a1c 6a6k 6agt 6ahs 6bqd 6cjj 6cjp 6cjr 6cjs 6cyg 6cyh 6d5w
6dyz 6dz0 6dz2 6dz3 6e3m 6e3n 6e3o 6e3p 6e3z 6e4c 6e6j 6e6v 6e6w 6fcj
6fe5 6ffe 6fff 6ffg 6g3c 6g5u 6gbw 6gga 6ggb 6ggd 6gge 6ggf 6gj5 6gj7
6gj8 6gwe 6h13 6h14 6hbn 6hhp 6hhr 6hlb 6hld 6hle 6hmt 6hmy 6hop 6hoq
6hor 6hot 6hou 6hza 6hzb 6hzc 6hzd 6i61 6i62 6i63 6i64 6i65 6i66 6i67
6i74 6i75 6i76 6i77 6i78 6i8m 6i8t 6ibx 6iby 6ibz 6ic0 6inz 6j06 6jmf
6jse 6jsf 6jsg 6jsn 6jsz 6jt3 6jut 6jwa 6k04 6k05 6k2n 6k3l 6miv 6miy
6mj4 6mja 6mji 6mjj 6mjq 6mo0 6mo2 6mo7 6mo9 6moa 6n4b 6n8x 6np2 6np3
6np4 6np5 6nrf 6nrg 6nrh 6nri 6nt2 6nv7 6nv9 6nw3 6o5u 6od6 6oi8 6oin
6oio 6oip 6oiq 6oir 6olx 6op9 6oxp 6oxq 6oxr 6oxs 6oxt 6oxu 6oxv 6oxw
6oxx 6oxy 6oxz 6oy0 6oy1 6oy2 6pyb 6pz4 6qfe 6qge 6qgf 6qi7 6qln 6qlo
6qlp 6qlq 6qlr 6qls 6qlt 6qlu 6qmr 6qmt 6qqt 6qqu 6qqv 6qqw 6qr0 6qr7
6qr9 6qra 6qrc 6quw 6qwi 6qxd 6rot 6rpg 6s55 6s56 6s57 6t6a 6ufn 6ufo
6uhv 6uii 6uim 6uvp 6uvv 6uvy 6uwp 6uwv 6v5l
###################################################################################

###################################################################################
# 24 extreme near-neighbor easy cases:
6a6k 6cyg 6cyh 6hza 6hzb 6i63 6iby 6oxp 6oxq 6oxr 6oxs 6oxt 6oxu 6oxv
6oxw 6oxx 6oxy 6oxz 6oy0 6oy1 6oy2 6qi7 6qlp 6qls
###################################################################################

\end{lstlisting}

\newpage
\section*{Appendix: Vina Docking Scripts}
Docking with Vina version 1.1.2 followed the procedure below (run within each of the Clean Test Set directories):
\begin{lstlisting}
###################################################################################
# Prepare receptor structures
prepare_receptor.bat -r protein.mol2 -U ignore -o p01_receptor.pdbqt

# Make a bounding box based on the cognate ligand --> known_box.txt
surflex-dock-v5189.exe bbox gold-lig.mol2 2.0 10.0 0 known

# Make the test ligand
prepare_ligand.bat -l lig-random.mol2 -o testlig-$1.pdbqt

# Run the docking with a fixed seed
vina.exe --receptor p01_receptor.pdbqt --ligand testlig-$1.pdbqt
         --config known_box.txt --exhaustiveness=32 --num_modes 20
         --seed -337841744
         --out p01_$1_vina_out.pdbqt >& p01_$1_vina-log

obabel p01_$2_vina_out.pdbqt -O p01_$1_vina_dock.mol2

# Cluster poses into families and generated automorph corrected RMSD values
# Assumes names of the ligand pose files will be as generated above
# Assumes Vina console output (which includes scores) will be as generated above
surflex-dock-v5189.exe rms_vina gold-lig.mol2 vinafam001 1
surflex-dock-v5189.exe rms_vina gold-lig.mol2 vinafam005 5
###################################################################################

\end{lstlisting}

\section*{Appendix: Gnina Docking Scripts}
Docking with Gnina version 1.1 (master:e4cb300+, Built 12-18-2023) followed the procedure below (run within each of the Clean Test Set directories) after the Vina docking procedure was run to produce the receptor files:
\begin{lstlisting}
###################################################################################
# Run gnina as recommended for cognate ligand re-docking:
gnina --cpu=2 -r p01_receptor.pdbqt -l lig-random.mol2
      --seed -337841744
      --num_modes=20 --exhaustiveness=32 --autobox_ligand gold-lig.mol2
      -o p01_$1_gnina_dock.mol2 >& p01_$1_gnina-log

# Cluster poses into families and generated automorph corrected RMSD values
# Assumes names of the ligand pose files will be as generated above
# Assumes Gnina console output (which includes scores) will be as generated above

surflex-dock-v5189.exe rms_gnina gold-lig.mol2 gninafam001 1
surflex-dock-v5189.exe rms_gnina gold-lig.mol2 gninafam005 5
###################################################################################

\end{lstlisting}

\newpage
\section*{Appendix: Glide Protein Preparation and Docking Scripts}
All results reported for Glide were generated using the following procedure:
\begin{enumerate}
  \item Prepare the proteins: \texttt{01\_protein\_prep.py} (note: original PDB file copied to \texttt{protein.pdb})

  \item Remove the ligands from the prepared proteins: \\ \texttt{\$SCHRODINGER/run 02\_extract\_ligand\_from\_mae\_protein.py}

  \item Build the Glide grids: \texttt{03\_build\_glide\_grid.py}

  \item Dock the molecules: \texttt{04\_run\_glide.py}

  \item Analyze the data: \texttt{05\_evaluate\_docking.ipynb}

\end{enumerate}

\texttt{01\_protein\_prep.py}
\begin{lstlisting}
#!/usr/bin/env python

from glob import glob
import os

def run_protein_prep(infile_name, outfile_name):
    cmd = "$SCHRODINGER/run /opt/schrodinger2022-3/mmshare-v5.9/python/scripts/prepwizard2_driver.py"
    options = "-fillsidechains -disulfides -rehtreat -max_states 1 -epik_pH 7.4 -epik_pHt 2.0 -antibody_cdr_scheme Kabat -samplewater -propka_pH 7.4 -f S-OPLS -rmsd 0.3 -watdist 5.0  -fix" 
    os.system(f"{cmd} {infile_name} {outfile_name} {options}")


def main():
    input_protein = "protein.pdb"
    prepped_protein = "protein_prepped_fix.maegz"

    filespec = "[5,6]*"
    for dirname in sorted(glob(filespec)):
        print(dirname)
        os.chdir(dirname)
        run_protein_prep(input_protein,prepped_protein)
        os.chdir("..")
        
main()    
\end{lstlisting}

\newpage
\texttt{02\_extract\_ligand\_from\_mae\_protein.py}
\begin{lstlisting}
#!/usr/bin/env python

from glob import glob
import os
from rdkit import Chem
from schrodinger import structure
from schrodinger.structutils import analyze
from schrodinger.test import mmshare_data_file
from schrodinger.protein.getpdb import download_pdb
from schrodinger.structutils import transform
import numpy as np

def check_atom_counts(dirname):

    os.chdir(dirname)

    ref_mol = Chem.MolFromMolFile("gold-lig.sdf",removeHs=False)
    ref_conf = ref_mol.GetConformer()
    pos = ref_conf.GetPositions()
    ref_center = np.mean(pos,axis=0)

    if os.path.isfile("protein_prepped_fix.maegz"):
        st = structure.StructureReader.read("protein_prepped_fix.maegz")
        found = False
        dist_list = []
        for mol in st.molecule:
            num_atoms = len(mol.atom)
            if num_atoms > 3 and num_atoms < 500:
                mol_center = np.mean(np.array([atm.xyz for atm in mol.atom]),axis=0)
                dist = np.linalg.norm(ref_center-mol_center)
                dist_list.append(dist)
                if dist < 1:
                    print(dirname,num_atoms,ref_mol.GetNumAtoms(),dist)
                    found = True
                    del_atm_list = [atm.index for atm in mol.atom]
        if found:
            st.deleteAtoms(del_atm_list)
            outfile_name = "protein_prepped_no_ligand_fix.mae"
            with structure.StructureWriter(outfile_name) as writer:
                writer.append(st)
        else:
            print(dirname,"FAIL",sorted(dist_list))

    
    os.chdir("..")
    

def main():
    filespec = "[5,6]*"
    for dirname in sorted(glob(filespec)):
        check_atom_counts(dirname)

main()    
\end{lstlisting}

\newpage
\texttt{03\_build\_glide\_grid.py}
\begin{lstlisting}
#!/usr/bin/env python

from glob import glob
import os
from rdkit import Chem
import useful_rdkit_utils as uru

def build_glide_grid(protein_file_name, ligand_file_name, grid_file_name):
    cmd = "$SCHRODINGER/glide"
    options = "-WAIT -OVERWRITE"
    mol = Chem.MolFromMolFile(ligand_file_name)
    center_x, center_y, center_z = uru.get_center(mol)
    tmplt_buff = f"""FORCEFIELD   OPLS_2005
GRID_CENTER   {center_x:.2f}, {center_y:.2f}, {center_z:.2f}
GRIDFILE   {grid_file_name}
INNERBOX   10, 10, 10
OUTERBOX   25, 25, 25
RECEP_FILE   {protein_file_name}
"""
    with open("glide_grid.in","w") as ifs:
        print(tmplt_buff, file=ifs)
    os.system(f"{cmd} glide_grid.in {options}")


def main():
    grid_protein = "protein_prepped_no_ligand_fix.mae"
    ref_ligand = "gold-lig.sdf"
    glide_grid = "glide-grid_fix.zip"

    filespec = "[5,6]*"
    for dirname in sorted(glob(filespec)):
        print(dirname)
        os.chdir(dirname)
        if os.path.exists(grid_protein):
            build_glide_grid(grid_protein,ref_ligand,glide_grid)
        os.chdir("..")
        
main()    
\end{lstlisting}

\newpage
\texttt{04\_run\_glide.py}
\begin{lstlisting}
#!/usr/bin/env python

from glob import glob
import os
from rdkit import Chem
import useful_rdkit_utils as uru

def run_glide(preped_ligand_file_name, grid_file_name):
    cmd = "$SCHRODINGER/glide"
    options = "-WAIT -OVERWRITE -adjust"
    tmplt_buff = f"""FORCEFIELD   OPLS_2005
GRIDFILE   {grid_file_name}
LIGANDFILE   {preped_ligand_file_name}
POSE_OUTTYPE   ligandlib_sd
POSES_PER_LIG   10
POSTDOCK_NPOSE   10
PRECISION   SP
"""
    with open("glide_dock.in","w") as ifs:
        print(tmplt_buff, file=ifs)
    os.system(f"{cmd} glide_dock.in {options}")
    glide_output_sdf = "glide_dock_lib.sdfgz"
    success = False
    if os.path.isfile(glide_output_sdf):
        success = True
        os.rename(glide_output_sdf,"glide_dock_lib_fix.sdf.gz")
        os.system("gunzip -f glide_dock_lib_fix.sdf.gz")
    return success


def main():
    prepped_ligands = "prepped_random_ligands.maegz"
    glide_grid = "glide-grid_fix.zip"
    
    filespec = "[5,6]*"
    for dirname in sorted(glob(filespec)):
        print(dirname)
        os.chdir(dirname)
        if os.path.exists(glide_grid) and os.path.exists(prepped_ligands):
            run_glide(prepped_ligands,glide_grid)
        os.chdir("..")
        
main()    
\end{lstlisting}

\noindent For the Jupyter Notebook \texttt{05\_evaluate\_docking.ipynb}, please refer to the Data Archive under \texttt{Glide\_Scripts}. Alternatively, the following command can be run following completion of the Glide docking script (using v5.190 or higher of the BioPharmics Platform docking module binary):

\texttt{sf-dock.exe rms\_list glide\_dock\_lib\_fix.sdf gold-lig.sdf 5 rmsglide}

The resulting \texttt{rmsglide-sum} file will contain automorph-corrected RMSD values for the top-1 and top-5 poses.

\newpage
\section*{Appendix: Data Archive}
The data archive can be found at https://www.jainlab.org/downloads/ (DiffDock Archive). The following is the (excerpted) README that describes the data and file structure of the archive:
\begin{lstlisting}

This archive contains the clean test cases from the DiffDock study
(290 complexes). It does not contain the full PDBBind_processed
complexes (roughly 18,000) due to size. The information here can be
used to replicate the Surflex-Dock primary results, to run another
docking method, or to visualize the Surflex-Dock results.

Comparison of DiffDock to fully automatic SF-Dock workflow (v5.173
Surflex Platform). Key scripts:

  RunGrind             Grab PDB structure and grind into mol2, checking
                       ligands against verifypdb.smi

  RunPrep              Figure out which ligand is the DiffDock one by 
                       checking if identical in topology --> gold-lig.mol2
                       (and protein.mol2)
                       Make cognate-ligand located protomol. Regenerate
                       memory-free coordinates of gold-lig.mol2
                       --> lig-random.mol2
                       Do -pgeom level ForceGen search on randomized ligand
                       --> pglig.sfdb

  RunDock              Run standard docking, pose family generation,
                       automorph-corrected RMSD calculation for
                       top 1 and 5 pose families.

  Glide_Scripts        Scripts for running the Glide protocol.

  ...

Key files:

  diffdock_rmsd_results_samp40.tab  Results from Hannes Stark on the
                                    samp40 DiffDock protocol

  TestNames                         All 363 DiffDock test cases

  ...


For docking with a new docking method, the only files really required
are gold-lig.mol2 (the xtal pose of the ligand), protein.mol2
(well-prepared protein structure), and lig-random.mol2 (a clean
memory-free starting point). Key files for PDBBind_processed/6A6K
(whose data are representative of the other cases):

    gold-lig.mol2               Well-processed ligand from grindpdb
    protein.mol2                Well-processed protein from grindpdb
    lig-random.mol2             3D regenerated memory-free starting conformation

    ...


\end{lstlisting}

\end{document}